\documentclass[sigconf, 9pt, authorversion]{acmart}

\pdfoutput=1

\usepackage[T1]{fontenc}
\usepackage[english]{babel}
\usepackage{textcomp}
\usepackage{xspace}
\usepackage{graphicx}
\usepackage{amsfonts}
\usepackage{amsmath}
\usepackage{booktabs}
\usepackage[acronym]{glossaries}
\usepackage[capitalize]{cleveref}
\usepackage{tikz}
\usepackage{multirow}
\usepackage{multicol}
\usepackage{colortbl}
\usepackage{siunitx}

\glsdisablehyper
\def\BibTeX{{\rm B\kern-.05em{\sc i\kern-.025em b}\kern-.08em
    T\kern-.1667em\lower.7ex\hbox{E}\kern-.125emX}}
    
\makeatletter
\newcommand\notsotiny{\@setfontsize\notsotiny\@vipt\@viipt}
\newcommand*{\toroman}[1]{\expandafter\@slowromancap\romannumeral #1@}
\makeatother
\newcommand\mperiod[1][\rlap]{#1{\;.}}
\newcommand\mcomma[1][\rlap]{#1{\;,}}

\def\shapes{3D Shapes\xspace}
\def\medic{MEDIC\xspace}
\def\faces{FACES\xspace}
\def\method{MTL-Split\xspace}

\makeatletter
\DeclareRobustCommand\onedot{\futurelet\@let@token\@onedot}
\def\@onedot{\ifx\@let@token.\else.\null\fi\xspace}
\def\eg{e.g\onedot} 
\def\ie{i.e\onedot} 
 
\def\etc{etc\onedot} 
 
\def\etal{\emph{et al}\onedot}
\makeatother

\definecolor{red}    {HTML}{b7211f}
\definecolor{orange} {HTML}{FFA500}
\definecolor{blue}   {HTML}{4169E3}
\definecolor{green}  {HTML}{147546}
\definecolor{purple} {HTML}{92268F}

\usetikzlibrary{math}
\usetikzlibrary{calc}
\usetikzlibrary{arrows}
\usetikzlibrary{arrows.meta}
\usetikzlibrary{shapes.arrows}
\usetikzlibrary{shapes.symbols}
\usetikzlibrary{shapes.geometric}
\usetikzlibrary{patterns}
\usetikzlibrary{patterns.meta}
\makeatletter
\tikzset{%
    >=stealth,
    ultra thin/.style= {line width=0.1pt},
    very thin/.style=  {line width=0.2pt},
    thin/.style=       {line width=0.4pt},
    semithick/.style=  {line width=0.6pt},
    thick/.style=      {line width=0.8pt},
    very thick/.style= {line width=1.2pt},
    ultra thick/.style={line width=1.6pt},
    fblue/.style={fill=blue!50},
    fred/.style={fill=red!50},
    forange/.style={fill=orange!50},
    fgreen/.style={fill=green!50},
    fgray/.style={fill=gray!50},
    lblue/.style={draw=blue},
    lred/.style={draw=red},
    lorange/.style={draw=orange},
    lgreen/.style={draw=green},
    lgray/.style={draw=gray},
    AN/.style={anchor=north},
    ANW/.style={anchor=north west},
    ANE/.style={anchor=north east},
    AE/.style={anchor=east},
    AW/.style={anchor=west},
    AS/.style={anchor=south},
    ASW/.style={anchor=south west},
    ASE/.style={anchor=south east},
    AC/.style={anchor=center},
	opaque node/.code 2 args={\tikzset{opacity=#1, text opacity=#2}},
	double color fill/.code 2 args={%
		\pgfdeclareverticalshading[%
		tikz@axis@top,tikz@axis@middle,tikz@axis@bottom%
		]{diagonalfill}{100bp}{%
			color(0bp)=(tikz@axis@bottom);%
			color(50bp)=(tikz@axis@bottom);%
			color(50bp)=(tikz@axis@middle);%
			color(50bp)=(tikz@axis@top);%
			color(100bp)=(tikz@axis@top)%
		}%
		\tikzset{%
			shade,%
			left color=#1,%
			right color=#2,%
			shading=diagonalfill%
		}%
	}%
}
\makeatother

\newcommand{\DatasetSize}{\ensuremath{K}}
\newcommand{\TaskSize}{\ensuremath{N}}
\newcommand{\Input}{\ensuremath{x}}
\newcommand{\GroundTruth}{\ensuremath{y}}
\newcommand{\LossFunction}{\ensuremath{\mathcal{L}}}
\newcommand{\Gradient}{\ensuremath{\nabla}}

\newcommand{\Backbone}{\ensuremath{M_{b}}}
\newcommand{\BackboneParameter}{\ensuremath{\psi}}
\newcommand{\BackboneOutput}{\ensuremath{Z_{b}}}
\newcommand{\BackboneLearningRate}{\ensuremath{\eta}}

\newcommand{\Task}{\ensuremath{H}}
\newcommand{\TaskParameter}{\ensuremath{\theta}}
\newcommand{\TaskOutput}{\ensuremath{\hat{y}}}
\newcommand{\TaskLearningRate}{\ensuremath{\alpha}}

\newacronym{sc}{SC}{Split Computing}
\newacronym{mtl}{MTL}{Multi-Task Learning}
\newacronym{stl}{STL}{Single-Task Learning}
\newacronym{dnn}{DNN}{Deep Neural Network}
\newacronym{loc}{LoC}{Local-only Computing}
\newacronym{roc}{RoC}{Remote-only Computing}
\newacronym{cnn}{ConvNet}{Convolutional Neural Network}
\newacronym{rnn}{RNN}{Recurrent Neural Network}
\newacronym{sgd}{SGD}{Stochastic Gradient Descent}
\newacronym{ml}{ML}{Machine Learning}
\newacronym{fc}{FC}{Fully Connected}
\newacronym{svm}{SVM}{Support Vector Machine}
\newacronym{relu}{ReLU}{Rectified Linear Activation Unit}
\newacronym{convnet}{ConvNet}{Convolutional Neural Network}
\newacronym{mlp}{MLP}{MultiLayer Perceptron}

\copyrightyear{2024}
\acmYear{2024}
\setcopyright{rightsretained}
\acmConference[DAC '24]{61st ACM/IEEE Design Automation Conference}{June 23--27, 2024}{San Francisco, CA, USA}
\acmBooktitle{61st ACM/IEEE Design Automation Conference (DAC '24), June 23--27, 2024, San Francisco, CA, USA}
\acmDOI{10.1145/3649329.3655686}
\acmISBN{979-8-4007-0601-1/24/06}

\begin{document}

\title[MTL-Split: Multi-Task Learning for Edge Devices using Split Computing]{MTL-Split: Multi-Task Learning for Edge Devices\\using Split Computing}

\author{
    Luigi Capogrosso\textsuperscript{1}, 
    Enrico Fraccaroli\textsuperscript{1,2}, 
    Samarjit Chakraborty\textsuperscript{2}, 
    Franco Fummi\textsuperscript{1}, 
    Marco Cristani\textsuperscript{1}
}
\affiliation{%
    \institution{%
        \textsuperscript{1}\texttt{name.surame@univr.it},\ %
        \textsuperscript{2}\texttt{enrifrac@cs.unc.edu}, \texttt{samarjit@cs.unc.edu}\\
        \textsuperscript{1}Department of Engineering for Innovation Medicine, University of Verona, Italy\\
        \textsuperscript{2}Department of Computer Science, The University of North Carolina at Chapel Hill, USA
    }
    \country{}
}

\renewcommand{\shortauthors}{Capogrosso, et al.}

\begin{abstract}
\gls{sc}, where a \gls{dnn} is intelligently split with a part of it deployed on an edge device and the rest on a remote server is emerging as a promising approach.
It allows the power of \glspl{dnn} to be leveraged for latency-sensitive applications that do not allow the entire \gls{dnn} to be deployed remotely, while not having sufficient computation bandwidth available locally.
In many such embedded systems scenarios, such as those in the automotive domain, computational resource constraints also necessitate \gls{mtl}, where the same \gls{dnn} is used for multiple inference tasks instead of having dedicated \glspl{dnn} for each task, which would need more computing bandwidth.
However, how to partition such a multi-tasking \gls{dnn} to be deployed within a \gls{sc} framework has not been sufficiently studied.
This paper studies this problem, and \method{}, our novel proposed architecture, shows encouraging results on both synthetic and real-world data.
The source code is available at \url{https://github.com/intelligolabs/MTL-Split}.
\end{abstract}


\maketitle

\glsresetall

\section{Introduction}
\label{sec:intro}

In the last decade, \glspl{dnn} have achieved state-of-the-art performance in various problems.
However, \gls{dnn} models often present computational requirements that cannot be met by most of the resource-constraint edge devices available today~\cite{capogrosso2024machine}.
This prohibits the full deployment of \gls{dnn}-based applications on these systems, leading to what is commonly known as the \gls{loc} approach.
However, using simplified models negatively affects the overall accuracy.
As such, the most common deployment approach of \gls{dnn}-based applications on resource-constraint edge devices is the \gls{roc}.
With this, the network runs on the server side, and the input is directly transferred from the edge device to the server through a network connection.
Then, the server computes the inferences and sends the output back to the device.
However, such data transfer could lead to excessive latency times, especially in degraded channel conditions.

As a compromise between the \gls{loc} and the \gls{roc} approaches, recently suggested \emph{\gls{sc}} frameworks~\cite{matsubara2022split} propose to split \gls{dnn} models into a head and a tail, deployed on edge device and server, respectively.
In particular, early implementations of \gls{sc}, like~\cite{kang2017neurosurgeon}, select a layer and divide the model to define the head and tail sub-models.
Instead, more recent \gls{sc} frameworks introduce the bottleneck concept to achieve in-model compression toward the global task~\cite{matsubara2019distilled}.

At the same time, current state-of-the-art approaches in different \gls{ml} applications rely on advanced learning procedures, such as the \emph{\gls{mtl}}~\cite{caruana1997multitask}.
In particular, \gls{mtl} is a paradigm in which multiple related tasks are jointly learned to improve the generalizability of a model by using shared knowledge across different aspects of the input.
This is achieved by jointly optimizing the model's parameters across all tasks, allowing the model to learn both task-specific and shared representations simultaneously.

\vspace*{-.2cm}
\paragraph{\textbf{Innovations}}
In this paper, we present a new combination of \gls{sc} and \gls{mtl} to solve multiple inference tasks on edge devices.
Solving multiple tasks with a common \glspl{dnn} can lead to substantial resource savings. 
For example, consider the automotive domain: detecting a person with a camera requires solving both a classification task (identifying pedestrians, vehicles, buildings, \etc{}) and a regression task (determining bounding boxes corresponding to the classifications).
Our proposed approach aims to solve multiple tasks simultaneously, \ie{}, $T_{1}\dots{}T_{N}$, where $N$ represents the number of tasks, using only a single neural network, in contrast to current methods, where the emphasis is on \gls{stl}, which would need $N$ neural networks to solve the tasks.

As a result, by employing \gls{mtl}, we enhance performance across multiple tasks, elevating the design challenge beyond that of preserving a single task's performance, as in regular \gls{sc}.
Further, this allows systems to operate effectively even in scenarios where data is scarce for specific tasks but abundant for others, as explained and theoretically demonstrated in~\cite{boursier2022trace}.
Lastly, the shared feature space output is remarkably lightweight, significantly reducing network latency encountered in \gls{sc} scenarios.

In summary, the main contributions of this paper are:
\begin{itemize}
\item A new \gls{sc} design merged with \gls{mtl}.
Our design handles multiple tasks concurrently, instead of the current focus on \gls{stl} in \gls{sc}. 
\item Through \gls{mtl}, we increase task performance, overcoming the challenge of preserving only the performance of the main task.
\item Finally, the output from the shared feature space is remarkably lightweight, significantly mitigating the impact of network latency in a \gls{sc} scenario.
\end{itemize}

\section{Related Work}
\label{sec:related}

This section provides an overview of distributed deep learning applications, specifically focusing on \gls{sc} and \gls{mtl}.

\subsection{Distributed deep learning}
We focus on architectures operating through a \gls{dnn} model $M(\cdot{})$, whose task is to produce the inference output $y$ from an input $x$.
Three types of paradigms used for distributed deep learning can be identified in the literature, \emph{viz.}, \gls{loc}, \gls{roc}, and \gls{sc}.

\paragraph{\textbf{Local-only Computing (LoC)}}
Under this policy, the entire computation is performed on the sensing devices.
Therefore, the edge device entirely executes the function $M(x)$.
Its advantage lies in offering low latency due to the proximity of the computing element to the sensor.
However, it may not be compatible with \gls{dnn}-based architectures that demand robust hardware capabilities.
Usually, simpler \gls{dnn} models $\bar{M}(x)$ that use specific architectures (\eg{}, depth-wise separable convolutions) are used to build lightweight networks, such as MobileNetV3~\cite{howard2019searching}.

Besides designing lightweight neural models, in the last few years, progress has been made in DNN compression.
Techniques such as network pruning and quantization~\cite{liang2021pruning}, or knowledge distillation~\cite{gou2021knowledge}, achieve an efficient representation of the neural network but with some quality degradation.

\paragraph{\textbf{Remote-only Computing (RoC)}}
The input $x$ is transferred through the communication network and processed at the remote system through the function $M(x)$.

This paradigm preserves full accuracy considering the higher computation bandwidth of the remote system but leads to high latency and communication bandwidth due to the input transfer, especially when the cloud is located far from the edge device.

\paragraph{\textbf{Split Computing (SC)}}
A typical \gls{sc} scenario is discussed in~\cite{eshratifar2019jointdnn}, where neither \gls{loc} nor \gls{roc} approaches are optimal, and a split configuration is an ideal solution.
The \gls{sc} paradigm divides the \gls{dnn} model into a head, executed by the local sensing device, and a tail, executed by the remote system.
It combines the advantages of both \gls{loc} and \gls{roc}, thanks to the lower latency and, more importantly, drastically reduces the required transmission bandwidth by compressing the input $x$ to be sent through the use of an autoencoder~\cite{matsubara2019distilled}.
We define the encoder and decoder models as $z_{l}=F(x)$ and $\bar{x}=G(z_{l})$, which are executed at the edge and remotely, respectively.
The distance $d(x,\bar{x})$ defines the performance of the encoding-decoding process.

One of the earliest works on \gls{sc} is the study by Kang~\etal{}~\cite{kang2017neurosurgeon}, in which the initial layers of a \gls{dnn} are the most suitable candidates for partitioning, as they optimize both latency and energy consumption.
Additionally, latency reduction can be achieved through two methods: quantization, as explored in~\cite{li2018auto}, and the utilization of lossy compression techniques prior to data transmission, as investigated in~\cite{choi2018deep}.
The concept of employing autoencoders to further compress the data to be transferred is discussed in various studies, such as~\cite{eshratifar2019bottlenet}.

Prevalent methods for identifying potential splitting points have evolved from architecture-based, to more refined neuron-based ones.
Within the domain of architecture-based approaches, in~\cite{sbai2021cut}, candidate split locations are where the size of the \gls{dnn} layers decreases. The rationale is that compressing information by autoencoders, where compression would still occur due to the shrinking of the architecture, seems reasonable.
On the other hand, in~\cite{cunico2022split} and~\cite{capogrosso2023split}, it was shown that not only the type of the layers but also the saliency of individual layers is a crucial factor.
A neuron's saliency is determined by its gradient in relation to the accurate decision.
Thus, optimal splitting points should be positioned following layers housing impactful neurons, to preserve the information flowing until then.

Notably, while existing approaches target \gls{stl} problems, we propose the first \gls{sc} solution for multi-task learning challenges.
Furthermore, while state-of-the-art methods strive to minimize the drop in accuracy, our approach aims at enhancing the accuracy.
To the best of our knowledge, the only work that combines concepts similar to ours is~\cite{zhang2022bandwidth}.
In this, the authors introduce task-oriented edge computing to reduce bandwidth consumption, which is different from our proposal.

\subsection{Multi-Task Learning (MTL)}
\gls{mtl} to solve multiple learning problems at the same time~\cite{caruana1997multitask}, can help us reduce inference time, improve accuracy, and increase data efficiency~\cite{standley2020tasks}.
In its basic formulation, \gls{mtl} uses a common representation to predict several outputs from a single input.
An important aspect of this procedure is the relationship between tasks and how much a shared representation can be transferred across tasks~\cite{zamir2018taskonomy}, or how to weight the losses of different tasks~\cite{kendall2018multi} to create a better joint optimization objective.

In recent years, numerous methods have emerged to address the simultaneous solving of multiple tasks, from approaches that learn how to weigh automatically the different tasks~\cite{gao2019nddr}, to more sophisticated transformer-based architectures~\cite{xu2022mtformer}.
Parallel work also has explored different theoretical aspects of \gls{mtl}, such as treating it as a multi-objective optimization~\cite{sener2018multi} or using game-theoretic optimizations~\cite{navon2022multi}. In particular, \gls{mtl} approaches have remained theoretical without practical implementation at the edge. Thus, our work addresses this gap. 

The \glspl{dnn} solving the different tasks in \gls{mtl} are commonly known as task-solving heads, which, we understand, might be confused with the head/tail terminology of \gls{sc}.
For the remainder of the paper, when we use the term head, we refer to the \gls{mtl} terminology (\ie{}, task-solving heads) and not the \gls{sc} one (\ie{}, head/tail).

\section{Methodology}
\label{sec:method}

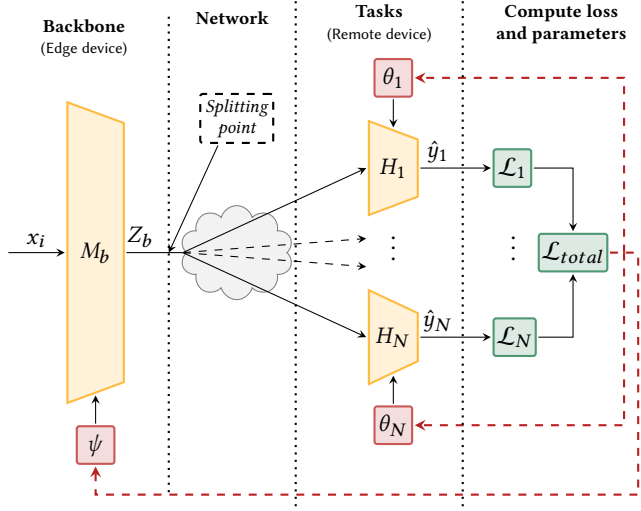
\begin{figure}[!tb]
    \centering
    \begin{tikzpicture}[
    every node/.append style={
        align = center, font=\normalsize, outer sep=1pt, inner sep=1pt
    },
    title/.append style={font=\footnotesize\bfseries},
    dnn/.append style={draw=orange!75, fill=orange!15},
    param/.append style={draw=red!75, fill=red!15},
    loss/.append style={draw=green!75, fill=green!15},
    trap0/.append style={
        draw, trapezium, thick, anchor=west,
        trapezium stretches body,
        trapezium angle=60,
        minimum width=4cm,
        minimum height=0.75cm,
        shape border rotate=270
    },
    trap1/.append style={
        trap0,
        trapezium angle=60,
        minimum width=1.25cm,
        minimum height=0.65cm
    },
    cloud0/.append style={
        cloud, AW, draw =gray, text =black, fill = gray!10,
        minimum width = 1.5cm, minimum height = 1.25cm
    },
    box0/.append style={
        draw, rectangle, thick, rounded corners = 1pt,
        minimum width  = 0.5cm, minimum height = 0.5cm
    }
]
\tikzmath{\scale=0.75;}
\draw[->] (0,0) --++ (1*\scale, 0) node[midway, above] {$\Input_{i}$}
    node[trap0, dnn] (bb) {$M_{b}$};
\draw[<-] (bb.south) --++ (0, -0.5*\scale) node[box0, param, AN] (bbp) {$\psi$};
\draw[-]  (bb.east) --++ (1*\scale, 0) node[pos=0.25, above] {$\BackboneOutput$}
    node[cloud0] (net) {} node[AC] at (net){};
\draw[->] (net.west) --++ (3.25*\scale, +1.5*\scale) node[trap1, dnn] (H1) {$H_{1}$};
\draw[<-] (H1.north) --++ (0, +0.5*\scale) node[box0, param, AS] (Hp1) {$\theta_{1}$};
\draw[->] (H1.east) --++ (1.25*\scale, 0) node[pos=0.25, above] {$\hat{y}_{1}$}
    node[box0, loss, AW] (L1) {$\mathcal{L}_{1}$};
\draw[->, dashed] (net.west) --++ (3.25*\scale, +0.25*\scale);
\draw[->, dashed] (net.west) --++ (3.25*\scale, -0.25*\scale);
\draw[->] (net.west) --++ (3.25*\scale, -1.5*\scale) node[trap1, dnn] (Hn) {$H_{N}$};
\draw[<-] (Hn.south) --++ (0, -0.5*\scale) node[box0, param, AN] (Hpn) {$\theta_{N}$};
\draw[->] (Hn.east) --++ (1.25*\scale, 0) node[pos=0.25, above] {$\hat{y}_{N}$}
    node[box0, loss, AW] (Ln) {$\mathcal{L}_{N}$};
\node[box0, loss, AW] (Ltot) at ($(L1.east)!0.5!(Ln.east)$) {$\mathcal{L}_{total}$};
\draw[->] (L1) -| (Ltot);
\draw[->] (Ln) -| (Ltot);
\draw[red, thick, dashed] (Ltot.east) --++(0.25*\scale,0) coordinate (cross1) --++(0.25*\scale,0) coordinate (cross2);
\draw[red, thick, dashed, ->] (cross1) |- (Hp1);
\draw[red, thick, dashed, ->] (cross1) |- (Hpn);
\draw[red, thick, dashed, <-] (bbp.south) --++(0, -0.5*\scale) -| (cross2);
\node[AC] at ($(H1.south)!0.35!(Hn.north)$) {\large\textbf\vdots};
\node[AC] at ($(L1.south)!0.4!(Ln.north)$) {\large\textbf\vdots};
\draw[dotted, thick, black]
    ($(bb.east)+(+.75*\scale,-4.25*\scale)$) --
    ($(bb.east)+(+.75*\scale,+4.25*\scale)$) coordinate (top1);
\draw[dotted, thick, black]
    ($(Hn.west)+(-1.25*\scale,-3*\scale)$) --
    ($(H1.west)+(-1.25*\scale,+3*\scale)$) coordinate (top2);
\draw[dotted, thick, black]
    ($(Ln.west)+(-.5*\scale,-3*\scale)$) --
    ($(L1.west)+(-.5*\scale,+3*\scale)$) coordinate (top3);
\node[AN, title] at ($(top1)-(1.5*\scale,0)$) {Backbone\\\scriptsize\normalfont\sffamily (Edge device)};
\node[AN, title] at ($(top1)!0.5!(top2)$) {Network};
\node[AN, title] at ($(top2)!0.5!(top3)$) {Tasks\\\scriptsize\normalfont\sffamily (Remote device)};
\node[AN, title] at ($(top3)+(1.75*\scale,0)$) {Compute loss\\and parameters};

\node[box0, dashed, font=\footnotesize\itshape, inner sep=2pt] (sp) at ($(net.north)+(0,1.5*\scale)$) {Splitting\\point};
\draw[->] ($(sp.south)-(0.25,0)$) -- ($(net.west)-(0.25*\scale, 0)$);
\end{tikzpicture}%
    \caption{The proposed architecture for handling complex inference tasks on edge devices by integrating \gls{sc} and \gls{mtl}.}
    \label{fig:architecture}
\end{figure}
This paper proposes to combine \gls{sc} and \gls{mtl} to execute complex inference tasks on edge devices.
After outlining our notation, this section delineates the formal components of our proposal, shown in \cref{fig:architecture}.
This architecture consists of two components:
\textit{i)} a shared \emph{backbone} deployed on the edge device, and \textit{ii)} a series of \emph{task-solving heads} on a single or multiple remote devices.
Orange trapezoids are \gls{dnn} models, while their parameters are enclosed in red boxes.
The green components on the right-hand are the loss functions used to update the learnable parameters.
A communication network separates edge and remote devices.

\paragraph{\textbf{Setting and notation}}
We assume the existence of a labeled image dataset defined as follows:
\begin{equation}\label{eq:dataset}
    D=\left\{(\Input_{i}, \GroundTruth_{i})\,|\,\forall{i}\in{}\{1\dots{}\DatasetSize\},\,x_{i}\in{}\mathbb{R}^{w\times{}h\times{}c},\, y_{i}\in{}\mathbb{N}^{\TaskSize}\right\} \mcomma
\end{equation}
where $\DatasetSize$ is the number of (image, labels) tuples, $\Input_{i}$ is the input representing the image, and $\GroundTruth_{i}$ a set of $\TaskSize$ labels associated with the $i$-th image, namely ground truth.
The input $\Input_{i}$ is a tensor with dimensions $w\times{}h\times{}c$, where $w$ is the width, $h$ is the height, and $c$ is the number of channels (\eg{}, red, green, blue).
In this work, we consider the \emph{classification task} that tries to learn a mapping from the image space $\{x_{i}|\forall{i}\in{}\{1\dots{}\DatasetSize\}\}$ to the corresponding set of labels $\{y_{i}|\forall{i}\in{}\{1\dots{}\DatasetSize\}\}$.

\subsection{Proposed architecture}
Unlike a classic \gls{sc} scenario, here we focus on architectures operating through a \gls{dnn} model, whose task is to produce the inference outputs $\{\TaskOutput_{j}|\forall{j}\in{}\{1\dots{}\TaskSize\}\}$ from an input $\Input_{i}$, where $\TaskSize$ is the number of tasks to be solved.
In this way, one can build a single model that learns multiple tasks across the same input.

As shown in \cref{fig:architecture}, the first module is the \textit{backbone}, a \gls{dnn} model $\Backbone(\cdot{})$ sharing hidden layers among all tasks.
In this way, we greatly reduce the risk of overfitting: the more tasks we learn concurrently, the more our model has to find a representation that captures all tasks, and the less likely we are to overfit the original task.
We describe the backbone operations on the $i$-th input as follows:
\begin{equation}\label{eq:backbone}
    \BackboneOutput = \Backbone(\Input_{i}; \BackboneParameter) \mcomma
\end{equation}
where $\BackboneParameter$ are the set of shared learnable parameters, and $\BackboneOutput$ is the backone's output.
The output $\BackboneOutput$ is typically a tensor, which, in our approach, is flattened before being sent through the network.
This output represents the point of the networks where the shared feature representation $\BackboneOutput$ is extracted from the backbone and transferred to the task-solving heads.

Each task is implemented by its own task-solving head, or head hereafter, a \gls{dnn} model $\Task_{j}$ located outside the edge devices, \eg{}, on a remote server.
We describe the operations of the $j$-th head $\Task_{j}$, as follows:
\begin{equation}\label{eq:task}
    \TaskOutput_{j} = \Task_{j}(\BackboneOutput; \TaskParameter_{j}) \mcomma
\end{equation}
where $\TaskParameter_{j}$ is its set of learnable parameters, and $\TaskOutput_{j}$ its output.

Putting it all together, the overall system output is a collection of outputs from all heads, organized either as a list or a single tensor.

\subsection{Training strategy}
The proposed methodology is architecture-independent.
Any neural network architecture can implement the backbone network and heads, such as a \gls{cnn} or a \gls{rnn}, designed to capture useful features from the input data $\Input_{i}$.
Regardless of the desired architecture, the objective of the \gls{mtl} system is to encourage the model to perform well on all tasks simultaneously.
Let us denote the \emph{task-specific loss function} for the $i$-th input and $j$-th task as $\LossFunction_{j}(\GroundTruth_{i}, \TaskOutput_{j})$, which measures the differences between corresponding label inside the ground truth $\GroundTruth_{i}$ and the predicted output $\TaskOutput_{j}$.
The \emph{overall loss function} for the \gls{mtl} system with the $i$-th input can be defined as the sum of losses from each task, as follows:
\begin{equation}\label{eq:total_loss}
    \LossFunction_{total} = \sum_{j = 1}^{\TaskSize} \LossFunction_{j}(\GroundTruth_{i}, \TaskOutput_{j})\mperiod
\end{equation}

The training process updates the shared backbone parameters $\BackboneParameter$ and the heads' parameters $\TaskParameter_{j}$ by backpropagating the gradient of the total loss with respect to these parameters and using an optimization algorithm like \gls{sgd}. 
The specific \gls{dnn} architecture, activation functions, and optimization methods can vary based on the problem and input data.

\subsection{Fine-tuning the model}
A key aspect of the proposed methodology, besides exploiting \gls{sc} to enable \gls{mtl} on edge devices, is the fine-tuning process explained in this section.
There are several reasons for performing fine-tuning, such as if we aim to enhance task-specific performance or if we want to introduce new tasks to the system.
During the fine-tuning phase, we update the heads' parameters $\TaskParameter_{j}$ while keeping the shared backbone parameters $\BackboneParameter$ relatively fixed.

During the fine-tuning process, heads' parameters are updated using gradients with respect to the task-specific loss:
\begin{equation}\label{eq:task_update}
    \TaskParameter_{j} := \TaskParameter_{j} - \TaskLearningRate \cdot \Gradient_{\TaskParameter_{j}}\LossFunction_{j}(\GroundTruth_{i}, \TaskOutput_{j}) \mcomma
\end{equation}
where $\TaskLearningRate$ is the learning rate for updating heads' parameters.
The shared backbone's parameters are often kept fixed or updated conservatively during fine-tuning.
As such, we need to define a separate update process as follows:
\begin{equation}\label{eq:backbone_update}
    \BackboneParameter := \BackboneParameter - \BackboneLearningRate \cdot \Gradient_{\BackboneParameter}\LossFunction_{total}\mcomma
\end{equation}
where $\BackboneLearningRate$ is the learning rate for updating the shared parameters, a small value compared to the one used to update the heads' parameters shown in \cref{eq:task_update}.

Given the parameters update functions, we can now define the fine-tuning process as an optimization problem, which involves minimizing the sum of the total loss as follows:
\begin{equation}
    \underset{\TaskParameter, \BackboneParameter}{\text{minimize}}\ \ \LossFunction_{total}\mperiod
\end{equation}

It is worth keeping in mind that fine-tuning should be approached with care.
It is important to find a balance between adapting the model to task-specific characteristics and retaining the general knowledge from the shared backbone. 
Too much fine-tuning can lead to overfitting on limited task-specific data, while too little fine-tuning might not fully harness the benefits of the \gls{mtl} setup.

\section{Experiments}
\label{sec:experiments}

This section describes the experimental trials that have been performed to validate our claims, along with their implementation details and results.

\paragraph{\textbf{Models details}}
In our experimental setup, we used three well-known \glspl{dnn}, \ie{}, VGG16~\cite{simonyan2014very}, MobileNetV3~\cite{howard2019searching}, and EfficientNet~\cite{tan2019efficientnet}, as a shared backbone.
We chose the first one because it is a well-established and widely used architecture in many image-processing tasks.
While the others represent cutting-edge \glspl{dnn} for embedded systems applications.
The task-solving heads are custom \gls{mlp} composed of two linear layers activated by the \gls{relu} function.
We want to point out that in this design, the task-solving heads are smaller than the backbone.
However, even though the task-solving heads are smaller than the backbone individually, if we consider a large number of tasks $N$, their combined size becomes larger than that of the backbone.
Due to this rationale, our architecture provides the deployment of task-solving heads on the remote server.

\paragraph{\textbf{Datasets}}
To effectively showcase the capabilities of our proposal, we begin our experiments with the \shapes{}~\cite{3dshapes18} dataset, a widely used toy benchmark in the \gls{ml} literature.
We further demonstrate the effectiveness of our proposed method by exploring its performance on the \medic{}~\cite{alam2023medic} and \faces{}~\cite{ebner2010faces} datasets, one of the well-known, and the newest \gls{mtl} benchmarks, respectively.

\shapes{} is a dataset of 3D shapes generated from 6 independent factors.
All possible combinations of these factors are present exactly once, resulting in 480,000 total images.
These are the floor hue, wall hue, object hue, scale, shape, and orientation.
Therefore, it is possible to treat the classification of each factor as a different task to solve, \ie{}, $T=[T_{1}\dots{}T_{6}]$.
Due to the straightforward nature of the synthetic images in \shapes{}, solving the classification tasks with a \gls{dnn} can be easy, leaving a limited possibility for improvement through \gls{mtl}.
Thus, to render this setting more realistic, we add salt-and-pepper noise of 15\% of the image pixels, making the classification more difficult.
In particular, with the presence of noise, the classification of object size (8 classes) and object type (4 classes) becomes challenging.

\medic{} is the largest social media image classification dataset for humanitarian response, consisting of 71,198 samples to address four different tasks.
Specifically, we decided to address only the damage severity (3 classes) and disaster type (4 classes) tasks since informativeness and humanitarian are somewhat trivial.

\faces{}, is a set of 2,052 images of naturalistic faces.
Here, the task corresponds to the classification of the perceived ages (3 classes), genders (2 classes), and facial expressions (3 classes).

\paragraph{\textbf{Training and inference details}}
All the code is implemented in PyTorch Lightning, and the used pre-trained network corresponds to the implementations in PyTorch~\cite{paszke2019pytorch}.
On the \shapes{} dataset, we train our models for 10 epochs, with a learning rate of $1\times{}10^{-5}$, using AdamW~\cite{loshchilov2017decoupled} as an optimizer, on a NVIDIA RTX 3090.
On the \medic{} and \faces{} dataset, we train our models for 50 epochs, with a learning rate of $1\times{}10^{-4}$, always using AdamW as an optimizer, on an NVIDIA RTX 3090.

We run all the experiments on an NVIDIA Jetson Nano with \SI{4}{\giga{}\byte{}} of memory.

\subsection{Multi-Task Learning (MTL) results}
In this section, we validate our claims within the \gls{mtl} context, which evidences the effectiveness of our proposal.
Given the distinct nature of our methodology (see Section~\ref{sec:related}), a direct comparison with state-of-the-art \gls{mtl} methods would be inadequate since these approaches are based on advanced loss functions~\cite{kendall2018multi} rather than models~\cite{xu2022mtformer}.
As a result, based on~\cite{caruana1997multitask}, our experimental protocol involves benchmarking our models against their respective single-task performance.

\begin{table}[t!]
    \centering
    \caption{Classification accuracy on the test partition of the \shapes{} dataset considering the object size ($T_{1}$) and the object type ($T_{2}$). 
    Values are reported as a percentage.}
    \begin{small}
    \begin{tabular}{ccc|cc}
\toprule
\multirow{2}{*}{Model} &
\multicolumn{2}{c}{\begin{tabular}{@{}c@{}}Single-Task Learning\end{tabular}} &
\multicolumn{2}{c}{\begin{tabular}{@{}c@{}}Multi-Task Learning\\($T_{1}$ + $T_{2}$)\end{tabular}} \\
\cmidrule(lr){2-3}\cmidrule(lr){4-5}
    & $T_{1}$ $\uparrow{}$ & $T_{2}$ $\uparrow{}$ & $T_{1}$ $\uparrow{}$ & $T_{2}$ $\uparrow{}$ \\ \midrule
\multicolumn{1}{l|}{VGG16}        &
12.50 & 25.50 & \textbf{51.10} \scriptsize{\color{green}{(+38.60)}} & \textbf{81.74} \scriptsize{\color{green}{(+56.24)}} \\
\multicolumn{1}{l|}{MobileNetV3}  &
74.85 & 93.95 & \textbf{77.23} \scriptsize{\color{green}{(~+2.38)}} & \textbf{94.00} \scriptsize{\color{green}{(~+0.05)}} \\
\multicolumn{1}{l|}{EfficientNet} &
95.49 & 99.07 & \textbf{96.66} \scriptsize{\color{green}{(~+1.17)}} & \textbf{99.48} \scriptsize{\color{green}{(~+2.28)}} \\
\bottomrule
\end{tabular}

    \end{small}
    \label{tab:shapes_results}
\end{table}
We begin our analysis on the \shapes{} dataset.
The results, in terms of accuracy, are shown in~\cref{tab:shapes_results} and demonstrate that \method{} improves the performance on all the tasks with respect to the \gls{stl} design choice.
This confirms the first two claims of our proposal, \ie{}, our architecture handles multiple tasks simultaneously and improves accuracy across the entire task set by collectively optimizing the model's parameters for all tasks.
Hence, in the context of \gls{sc} (which encompasses multiple tasks to be solved), our approach guarantees performance improvement rather than merely aiming to minimize performance degradation, which is the \gls{sc} trend observed in all previous state-of-the-art methods. 
Furthermore, the ability to address multiple tasks within the same network simultaneously has resulted in space and computational savings during inference because it only requires the evaluation of a single network instead of $N$ neural networks to solve each task.

\begin{table}[t!]
    \centering
    \caption{Classification accuracy on the test set of the \medic{} dataset considering the damage severity ($T_{1}$), and disaster type ($T_{2}$).
    Values are reported as a percentage.}
    \begin{small}
    \begin{tabular}{ccc|cc}
\toprule
\multirow{2}{*}{Model} &
\multicolumn{2}{c}{\begin{tabular}{@{}c@{}}Single Task Learning\end{tabular}} &
\multicolumn{2}{c}{\begin{tabular}{@{}c@{}}Multi Task Learning\\($T_{1}$ + $T_{2}$)\end{tabular}} \\
\cmidrule(lr){2-3}\cmidrule(lr){4-5}
    & $T_{1}$ $\uparrow{}$ & $T_{2}$ $\uparrow{}$ & $T_{1}$ $\uparrow{}$ & $T_{2}$ $\uparrow{}$ \\ \midrule
\multicolumn{1}{l|}{VGG16}        &
61.78 & 59.14 & \textbf{62.65} \scriptsize{\color{green}{(+0.87)}} & \textbf{60.54} \scriptsize{\color{green}{(+1.40)}} \\
\multicolumn{1}{l|}{MobileNetV3}  &
61.73 & 52.66 & \textbf{61.90} \scriptsize{\color{green}{(+0.17)}} & 52.29 \scriptsize{\color{red}{(-0.37)}}            \\
\multicolumn{1}{l|}{EfficientNet} &
61.00 & 53.94 & \textbf{62.42} \scriptsize{\color{green}{(+1.42)}} & \textbf{55.74} \scriptsize{\color{green}{(+1.80)}} \\
\bottomrule
\end{tabular}
    
    \end{small}
    \label{tab:medic_results}
\end{table}
\cref{tab:medic_results} further demonstrates the efficacy of our proposal exploring its performance on the \medic{} dataset.
This experiment serves as a compelling validation of our previous claims, showcasing the robustness of our architecture even when applied to complex datasets. 
In this case, it is important to highlight the \textit{inductive transfer} between tasks, as even a small increase in decimal points in this challenging context represents a significant achievement.

The minor decrease of 0.37\% of $T_{2}$'s performance in the \gls{mtl} setting does not represent a problem.
Specifically, what is known in the \gls{mtl} literature as \textit{negative transfer} comes up when there is a significant deterioration in performance across all tasks, typically resulting from conflicting or unrelated task objectives.
Since the performance improves in all the other cases, we can confidently say that negative transfer doesn't occur here.
We attribute this outcome to gradient fluctuations.
These results demonstrate the effectiveness of our approach, as it consistently yields significant improvements even in difficult scenarios.

\begin{table*}[t!]
    \centering
    \caption{Classification accuracy on the test set of the \faces{} dataset considering the the perceived ages ($T_{1}$), genders ($T_{2}$), and facial expressions ($T_{3}$). 
    Values are reported as a percentage.}
    \begin{small}
    \begin{tabular}{cccc|cc|cc|ccc}
\toprule
\multirow{2}{*}{Model} &
\multicolumn{3}{c}{\begin{tabular}{@{}c@{}}Single Task Learning\end{tabular}} &
\multicolumn{2}{c}{\begin{tabular}{@{}c@{}}Multi Task Learning\\($T_{1}$ + $T_{3}$)\end{tabular}} &
\multicolumn{2}{c}{\begin{tabular}{@{}c@{}}Multi Task Learning\\($T_{2}$ + $T_{3}$)\end{tabular}} &
\multicolumn{3}{c}{\begin{tabular}{@{}c@{}}Multi Task Learning\\($T_{1}$ + $T_{2}$ + $T_{3}$)\end{tabular}}\\
\cmidrule(lr){2-4}\cmidrule(lr){5-6}\cmidrule(lr){7-8}\cmidrule(lr){9-11}
                                  & $T_{1}$ $\uparrow{}$ & $T_{2}$ $\uparrow{}$ & $T_{3}$ $\uparrow{}$ & $T_{1}$ $\uparrow{}$ & $T_{3}$ $\uparrow{}$ & $T_{2}$ $\uparrow{}$ & $T_{3}$ $\uparrow{}$ & $T_{1}$ $\uparrow{}$ & $T_{2}$ $\uparrow{}$ & $T_{3}$ $\uparrow{}$ \\ \midrule
\multicolumn{1}{l|}{VGG16}        &
96.83 & 95.61 & 19.02 & 
\textbf{97.80} \scriptsize{\color{green}{(+0.87)}} & \textbf{91.46} \scriptsize{\color{green}{(+72.44)}} & 
\textbf{99.02} \scriptsize{\color{green}{(+3.41)}} & \textbf{90.24} \scriptsize{\color{green}{(+80.22)}} & 
\textbf{98.54} \scriptsize{\color{green}{(+1.71)}} & \textbf{99.51} \scriptsize{\color{green}{(+3.90)}}  & \textbf{89.27} \scriptsize{\color{green}{(+70.25)}} \\
\multicolumn{1}{l|}{MobileNetV3}
    & 97.07 & 99.51 & 95.12 &
    \textbf{99.51} \scriptsize{\color{green}{(+2.44)}} &
    95.12 \scriptsize{(+0.00)} & 99.51 \scriptsize{(+0.00)} &
    \textbf{95.61} \scriptsize{\color{green}{(+0.49)}} & 
    \textbf{99.27} \scriptsize{\color{green}{(+2.20)}} & 99.51 \scriptsize{(+0.00)} & \textbf{95.85} \scriptsize{\color{green}{(+0.73)}}\\
\multicolumn{1}{l|}{EfficientNet} &
    99.76 & 99.76 & 94.63 & 
    \textbf{100} \scriptsize{\color{green}{(+0.24)}} & \textbf{95.61} \scriptsize{\color{green}{(+0.98)}} & 
    99.76 \scriptsize{(+0.00)} & \textbf{97.32} \scriptsize{\color{green}{(+2.96)}} & 
    \textbf{100} \scriptsize{\color{green}{(+0.24)}} & \textbf{100} \scriptsize{\color{green}{(+0.24)}}   & \textbf{95.61} \scriptsize{\color{green}{(+0.98)}} \\
\bottomrule
\end{tabular}

    \end{small}
    \vspace*{-0.2em}
    \label{tab:faces_results}
\end{table*}
Finally, \cref{tab:faces_results} shows the results achieved on the \faces{} dataset employing the fine-tuning strategy starting from pre-trained networks on ImageNet.
The overall accuracies obtained are quite high, which was expected given the utilization of a pre-trained network as a starting point.
However, once again, our approach demonstrated its efficacy in enhancing performance across all tasks.
This is significant since it increases accuracies approaching near-maximum values.
Usually, such improvements necessitate the network's ability to correctly classify the intricate corner cases within the datasets.
In all instances where we do not achieve performance improvements, our results consistently align with the single-task performance (also in this case, ruling out the possibility that the non-performance improvements are due to negative transfer).

\subsection{Split Computing (SC) analysis}
\begin{table*}[t!]
    \centering
    \caption{Computing the size of the backbone \Backbone, and of its output \BackboneOutput.
    The reader should pay particular attention to the green columns in the table, as these are the columns displaying the results, which show that our proposal is really efficient for \gls{sc}.}
    \begin{footnotesize}
    \begin{tabular}{c|c|c|c|>{\columncolor{green!10}}c|c|>{\columncolor{green!10}}c}
\toprule
Model                              & 
\Backbone{} \#params (M)           & 
\Backbone{} \#params size (MB)     & 
Forward/backward pass size (MB)    & 
\Backbone{} estimated size (MB)    & 
\BackboneOutput{} \#params (M)     & 
\BackboneOutput{} size (MB)        \\
\midrule
MobileNetV3     & 0.9   & 3.58  & 724.08    & 727.66    & 55.3      & 0.21 \\
EfficientNet    & 4     & 15.45 & 3452.09   & 3467.54   & 406.06    & 1.56 \\
\bottomrule
\end{tabular}

    \end{footnotesize}
    \label{tab:sc_results}
\end{table*}

In this section, we examine the advantages of our approach in comparison to the other types of distributed deep learning paradigms while also presenting deployment considerations.

\paragraph{\textbf{Local-only Computing (LoC)}}
Under this paradigm, for the \shapes{} and the \medic{}, two distinct \glspl{dnn} are required since we address two different tasks.
Hence, the estimated memory size utilizing MobileNetV3 as a backbone is $\approx{}\SI{1.5}{\giga{}\byte{}}$, while it is $\approx{}\SI{6.9}{\giga{}\byte{}}$ for the EfficientNet.
Instread, for \faces{}, which involves three different tasks (\ie{}, three distinct \glspl{dnn} are required), the estimated memory size using MobileNetV3 is $\approx{}\SI{2.1}{\giga{}\byte{}}$, and for the EfficientNet is $\approx{}\SI{10.3}{\giga{}\byte{}}$.

As a result, due to memory constraints, the only feasible implementation on the Jetson Nano is restricted to MobileNetV3 on the \shapes{} dataset.
However, as indicated in \cref{tab:sc_results}, our approach, utilizing a single shared backbone on the edge device, enables the execution of all implementations on the same board.
Specifically, using  EfficientNet, we achieve memory size improvements of $\approx{}38\%$ for the \shapes{} and \medic{} datasets, and $\approx{}57\%$ for the \faces{} dataset.
As aforementioned, VGG16 is not optimal for embedded system applications, so we do not report data on that model.

\paragraph{\textbf{Remote-only Computing (RoC)}}
Under this policy, the goal is to minimize the data sent from the backbone to task-solving heads.

In the \faces{} dataset, the images are RGB with  $2835\times{}3543$ pixels.
Consequently, transmitting each input from the edge to the cloud involves transefing a tensor of size $2835\times{}3543\times{}3$, equivalent to $\approx{}\SI{115}{\mega{}\byte{}}$ over the network channel.

Whereas,~\cref{tab:sc_results} also highlights the minimal burden placed on the network channel when employing our methodology, thanks to the shared backbone's neural processing.
For example, assuming a gigabit channel: the time required to transfer 100 inputs of size $\approx{}\SI{115}{\mega{}\byte{}}$ each is $\approx{}\SI{98}{\second{}}$, whereas for our inputs of the size of $\SI{1.5}{\mega{}\byte{}}$, it is $\approx{}\SI{12}{\second{}}$, \ie{}, we obtain an improvement of $\approx{}87\%$ in the overall latency time.
This is important as Internet congestion will increasingly be driven by machine learning workloads.

This claim holds significant importance in a world that increasingly relies on efficient data transmission and reduced network congestion.

\vspace*{-0.3em}
\paragraph{\textbf{Discussion}}
The above analyses showcase the advantages of our proposal compared to \gls{loc}.
Our approach also excels in terms of data transmission, resulting in reduced total latency compared to \gls{roc}.
Furthermore, our design handles multiple tasks concurrently, thereby enhancing overall accuracy across all tasks.

\section{Concluding Remarks}
\label{sec:conclusions}

In this paper, we propose a new combination of \gls{sc} and \gls{mtl} to execute complex inference tasks on embedded devices.
The proposed architecture consists of a unified backbone that serves all tasks at the edge, and multiple task-solving heads situated outside the edge device.
This design enables the architecture to tackle multiple tasks simultaneously, in contrast to \gls{sc} methodologies that focus solely on a single task.
Furthermore, by incorporating \gls{mtl}, our approach enhances the accuracy performance across all tasks.
Additionally, the output from the shared backbone is notably lightweight, significantly mitigating network latency's impact in the context of \gls{sc}.
Extensive experimental validation confirms our claims.
Our approach succeeds at delivering superior performance across all tasks, whether they are simple or complex.
Further, comparison with other types of distributed deep learning paradigms and the deployment considerations confirm the efficacy of our design.

\balance

\begin{acks}
This study was carried out within the PNRR research activities of the consortium iNEST (Interconnected North-Est Innovation Ecosystem) funded by the European Union Next-GenerationEU (Piano Nazionale di Ripresa e Resilienza (PNRR) – Missione 4 Componente 2, Investimento 1.5 – D.D. 1058  23/06/2022, ECS\_00000043), and by the European Union’s Horizon Europe research and innovation programme under the Marie Sklodowska-Curie grant agreement No. 101109243.
This manuscript reflects only the Authors’ views and opinions. Neither the European Union nor the European Commission can be considered responsible for them.
This work was also partially supported by the US NSF grant 2038960. 
\end{acks}

\bibliographystyle{ACM-Reference-Format}
\bibliography{bibliography}

\end{document}